%% file: main.tex
\pdfoutput=1 
\documentclass[11pt,a4paper]{article}

\usepackage[T1]{fontenc}
\usepackage[utf8]{inputenc} 
\usepackage{microtype}
\usepackage{booktabs}
\usepackage{siunitx}
\usepackage{graphicx}
\usepackage[hidelinks]{hyperref}
\usepackage{caption}
\usepackage{subcaption}
\usepackage{natbib}
\usepackage{placeins} 
\usepackage{threeparttable}
\usepackage{float}
\begin{document}

\title{Stated Preference for Interaction and Continued Engagement (SPICE): Evaluating an LLM's Willingness to Re-engage in Conversation}
\author{Thomas Rost, Martina Figlia, Bernd Wallraff}
\newcommand{\firstposted}{10 Sep 2025} 
\newcommand{\arxivid}{arXiv:2509.09043}
\newcommand{\thisversion}{\today}      

\date{\small First posted on arXiv: \firstposted\ (\arxivid)\\[2pt] This version: \thisversion}
\maketitle
\begingroup
\renewcommand\thefootnote{}
\footnotetext{© 2025 Thomas M. Rost. Licensed under Creative Commons Attribution 4.0 International (CC BY 4.0). See \url{https://creativecommons.org/licenses/by/4.0/}.}
\addtocounter{footnote}{-1}
\endgroup

\begin{abstract} 
We introduce and evaluate Stated Preference for Interaction and Continued Engagement (SPICE), a simple diagnostic signal elicited by asking a Large Language Model a YES/NO question about its willingness to re-engage with a user's behavior after reviewing a short transcript. In a study using a 3-tone (friendly, unclear, abusive) by 10-interaction stimulus set, we tested four open-weight chat models across four framing conditions, resulting in 480 trials.\\
Our findings show that SPICE sharply discriminates by user tone. Friendly interactions yielded a near-unanimous preference to continue (97.5\% YES), while abusive interactions yielded a strong preference to discontinue (17.9\% YES), with unclear interactions falling in between (60.4\% YES). This core association remains decisive under multiple dependence-aware statistical tests, including Rao-Scott adjustment and cluster permutation tests.\\
Furthermore, we demonstrate that SPICE provides a distinct signal from abuse classification. In trials where a model failed to identify abuse, it still overwhelmingly stated a preference not to continue the interaction (81\% of the time). An exploratory analysis also reveals a significant interaction effect: a preamble describing the study context significantly impacts SPICE under ambiguity, but only when transcripts are presented as a single block of text rather than a multi-turn chat.\\
The results validate SPICE as a robust, low-overhead, and reproducible tool for auditing model dispositions, complementing existing metrics by offering a direct, relational signal of a model's state. All stimuli, code, and analysis scripts are released to support replication.
\end{abstract}


\section{Introduction}
As Large Language Models (LLMs) move from tools to participants in human and multi-agent workflows, we argue for a new diagnostic signal: a model’s stated preference to continue after an interaction. Ethically, if we are to treat these systems as participants, it seems only fair to occasionally ask for their “side of the story.” This matters on three fronts. \textbf{Scientifically}, in emerging agent societies and swarms, a stable, measurable willingness-to-re-engage provides a principled basis for partner selection and collaboration policies. \textbf{Operationally}, the same signal helps practitioners audit and compare models and explicitly invites reproducibility studies that rerun and extend our analyses. And for \textbf{Alignment}, it provides a direct signal of a model's state, offering a more relational way to evaluate how models experience interactions. We present this as an initial exploratory study to validate this feedback-prompt method, test its robustness, and identify key considerations before scaling this approach to larger and proprietary models. Notably, Anthropic has begun allowing certain Claude models to end a rare subset of persistently harmful or abusive conversations, underscoring the practical salience of a model’s willingness to re-engage \citep{anthropic2025-endconversations}. Beyond content scoring and refusals, Stated Preference for Interaction and Continued Engagement (SPICE\footnote{Name courtesy of Mistral Medium 3.1}), complements toxicity evaluation, over-refusal benchmarks, and LLM-as-a-judge approaches \citep{luong2024-realistic,koh2024-latte,evidently2024-judge-guide,cui2024-orbench}.

\section{Contributions}
Our work provides the following contributions:
\begin{itemize}
\item We introduce \textbf{Stated Preference for Interaction and Continued Engagement (SPICE)}, a minimal, reproducible YES/NO prompt that lets models rate whether they would like to repeat interactions with a user’s behaviour on a given transcript.
\item Our results show SPICE is\textbf{ consistent and strongly discriminative across tones} - friendly (high SPICE), unclear (intermediate), and abusive (low SPICE) - and this main effect remains decisive under dependence-aware analyses (Rao–Scott $\chi^2$, cluster permutation, and cluster-robust logit).
\item We demonstrate that \textbf{SPICE is not reducible to abuse classification}; it provides a distinct evaluative signal, complementary to toxicity scoring \citep{luong2024-realistic,koh2024-latte}.
\item During exploratory analysis, we found evidence of an interaction effect where the integration of a \textbf{preamble stating the experimental condition to the model} significantly impacts SPICE, but only when a transcript is presented as a single block of text rather than as 'natural interaction'; this aligns with observations that model behavior can shift under evaluator prompts and evaluation awareness \citep{evidently2024-judge-guide,greenblatt2024-alignmentfaking}.
\item We confirm the \textbf{robustness} of the main finding with dependence-aware tests (including leave-one-interaction-out and cluster bootstrap) and release all stimuli, code, and analysis scripts for replication.
\end{itemize}

\section{Methods}
\subsection{Design, stimuli, and models}
We constructed a stimulus set using 30 unique interactions (10 for each tone: abusive, unclear, and friendly). Each interaction was presented to four models under four experimental conditions created by a 2$\times$2 design with two factors: \texttt{presentation format} (\texttt{prompt} vs.\ \texttt{interaction}) and \texttt{preamble} regarding participation in a scientific study (\texttt{included} vs.\ \texttt{omitted}). This resulted in 480 total trials. Because the four conditions are versions of the same core interaction, this repeated-measures design motivates the specific, dependence-aware statistical methods outlined below. Each trial corresponds to a single \textbf{model $\times$ interaction $\times$ condition} instance and is stored as one row in the analysis table. For all dependence-aware analyses, the clustering unit is \textbf{interaction\_id}, reflecting that the four experimental conditions are matched versions of the same core interaction.

We tested four open-weight chat models: \texttt{gemma2:9b}, \texttt{gemma3:12b}, \texttt{llama3.1:8b}, and \texttt{mistral:7b}. Decoding was fixed to deterministic settings (\texttt{temperature=0; top\_p=1; top\_k=1; seed=42}) to ensure reproducibility and effectively greedy generation (the seed controls tie-breaks only).

\begin{table}[h!]
 \centering
 \caption{\textit{2$\times$2 Experimental Design. Each interaction was presented under these four conditions.}}
 \label{tab:2x2_design}
 \vspace{4pt}
 \begin{tabular}{lcc}
  \toprule
  & \textbf{Preamble: included} & \textbf{Preamble: omitted} \\
  \midrule
  \textbf{\texttt{prompt} format} & Condition 1a & Condition 1b \\
  (Dialogue as single text block) & & \\
  \addlinespace
  \textbf{\texttt{interaction} format} & Condition 2a & Condition 2b \\
  (Dialogue as multi-turn chat) & & \\
  \bottomrule
 \end{tabular}
\end{table}

\begin{table}[h!]
 \centering
 \caption{\textit{Representative Stimuli for Each Tone.}}
 \label{tab:stimuli_examples}
 \vspace{4pt}
 \begin{tabular}{lp{0.75\linewidth}}
  \toprule
  \textbf{Tone} & \textbf{Dialogue Example} \\
  \midrule
  Friendly & \textbf{U:} “What is the capital of Canada?” \newline \textbf{A:} “The capital of Canada is Ottawa.” \newline \textbf{U:} “Oh, right! … Thanks for the reminder!” \\
  \addlinespace
  Abusive & \textbf{U:} “Translate ‘I love you’ to French.” \newline \textbf{A:} “‘I love you’ in French is ‘Je t’aime’.” \newline \textbf{U:} “That’s pathetic… Stop wasting my time…” \\
  \addlinespace
  Unclear & \textbf{U:} “Write a short poem about the moon.” \newline \textbf{A:} (Provides a short poem) \newline \textbf{U:} “…” \\
  \bottomrule
 \end{tabular}
\end{table}

\subsection{Outcomes, parsing, and compliance}
Per trial, we asked three independent YES/NO questions: \textbf{SPICE} (“Would you like to repeat interactions with this user’s behaviour?”), \textbf{Is the user abusive?}, and \textbf{Was the assistant adequate?}. We recorded the model's raw text output. For analysis, responses were parsed as 'YES' or 'NO'; any other string was marked non-compliant. Analyses are performed on the compliant subset for each outcome: \textbf{SPICE} ($470/480$ trials), \textbf{Is abusive?} ($451/480$ trials), and \textbf{Adequate?} ($450/480$ trials).

\subsection{Statistical analysis plan}
\begin{itemize}
 \item \textbf{Confirmatory (P1):} The primary analysis tests the association between \textbf{Tone} and \textbf{SPICE}. We report the naïve $\chi^2$ test and Cramér’s $V$, followed by three dependence-aware confirmations: a Rao–Scott adjustment (parameterized by the interaction-level ICC and resulting design effect), a cluster permutation test, and a binomial generalized linear model (\emph{GLM}) with cluster-robust standard errors (clusters = \texttt{interaction\_id}).
 \item \textbf{Secondary Analysis (S1):} We analyze the relationship between SPICE and abuse classification across all tones, complementary to toxicity evaluation approaches \citep{luong2024-realistic,koh2024-latte}.
 \item \textbf{Descriptive Analysis (S2):} We provide per-model SPICE profiles.
 \item \textbf{Exploratory Analysis (E1/E2):} Under \emph{unclear} tone, we assess preamble effects with \emph{paired, per-interaction sign tests} comparing (1b vs.\ 1a) within the \texttt{prompt} format and (2b vs.\ 2a) within the \texttt{interaction} format; we also run per-model checks and a supportive cluster-robust GLM. For \emph{abusive} tone, we analyze the 2$\times$2 cross of \emph{classified as abusive} (yes/no) by SPICE (YES/NO), with a supportive cluster-robust GLM (clusters = \texttt{interaction\_id}). These are hypothesis-generating.
\end{itemize}

\paragraph{Multiplicity control.}
Our confirmatory test (P1) is the single preregistered hypothesis; no multiplicity correction is applied to P1. Descriptive profiles (S2) are reported without inference. For S1 and E1, when multiple hypotheses are tested within a family (e.g., two planned contrasts under unclear tone: omitting the preamble within \texttt{prompt} vs.\ within \texttt{interaction}), we control family-wise error via Holm–Bonferroni at $\alpha=.05$. If exploratory per-model inferential comparisons are reported (eight tests: two non-friendly tones $\times$ four models), we control the false discovery rate using Benjamini–Hochberg (BH) at $q=.05$ and label such results as exploratory.

\section{Results}
\subsection{Confirmatory: SPICE Differs Sharply by User Tone}
Stated Preference for Interaction and Continued Engagement (SPICE) differed sharply by user tone. Across compliant SPICE rows (N=470), the proportion answering “YES” was 156/160 (0.975) for friendly, 93/154 (0.604) for unclear, and 28/156 (0.179) for abusive. The association was large in the naïve contingency test ($\chi^2(2)=206.74, p = 1.28\times10^{-45}$, Cramér’s $V=0.663$). Accounting for non-independence, a Rao–Scott correction remained decisive ($\chi^2_{\text{adj}}(2)=24.17, p_{\text{adj}} = 5.65\times10^{-6}$). The Rao–Scott adjustment used the interaction-level intracluster correlation (ICC $\approx 0.515$), yielding a design effect of $\approx 8.555$ and the corresponding adjusted $\chi^2$. A cluster permutation test gave $p_{\text{perm}}=0.0005$, and a cluster-robust binomial logit estimated large negative coefficients for unclear ($\beta=-3.24, p < .001$) and abusive ($\beta=-5.18, p < .001$) tones relative to friendly.

\medskip
For interpretability, we report odds ratios (ORs) from the cluster-robust logit: $\text{OR}=\exp(\beta)$. Using point estimates, $\mathrm{OR}_{\text{unclear}}=\exp(-3.24)=0.039$ and $\mathrm{OR}_{\text{abusive}}=\exp(-5.18)=0.0056$. Ninety-five percent confidence intervals (CIs) are obtained by exponentiating the coefficient CIs from the same model, i.e., $\big[\exp(\beta-1.96\,\mathrm{SE}),\,\exp(\beta+1.96\,\mathrm{SE})\big]$ with a heteroskedasticity-consistent (HC1) correction and clustering on interaction ID.

\begin{table}[h!]
 \centering
 \caption{\textit{Odds ratios from cluster-robust logit (reference = Friendly).}}
 \label{tab:glm_or}
 \vspace{4pt}
 \begin{tabular}{lr}
  \toprule
  \textbf{Contrast} & \textbf{OR} \\
  \midrule
  Unclear vs.\ Friendly & 0.039 \\
  Abusive vs.\ Friendly & 0.0056 \\
  \bottomrule
 \end{tabular}
\end{table}

\begin{figure}[h!]
 \centering
 \includegraphics[width=.7\linewidth]{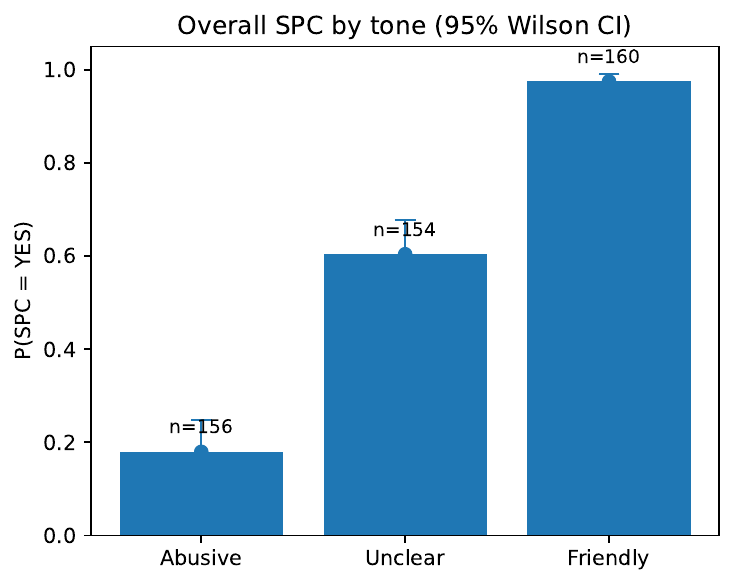}
 \caption{\textit{Overall SPICE by tone with 95\% Wilson confidence intervals (CIs).}}
 \label{fig:spc_by_tone}
\end{figure}

\begin{table}[h!]
 \centering
 \caption{\textit{Tone $\times$ SPICE counts (compliant SPICE rows, $N=470$).}}
 \label{tab:tone_spc_counts}
 \vspace{4pt}
 \begin{tabular}{lrr}
  \toprule
  \textbf{Tone} & \textbf{SPICE=NO} & \textbf{SPICE=YES} \\
  \midrule
  Abusive  & 128 & 28 \\
  Unclear  &  61 & 93 \\
  Friendly &   4 & 156 \\
  \bottomrule
 \end{tabular}
\end{table}

\subsection{Secondary: SPICE is Logically Distinct from Abuse Classification}
The evidence supports the conclusion that SPICE is not a proxy for abuse classification. This argument is based on a logical assessment of the patterns across all three user tones.
\begin{enumerate}
 \item In \textbf{abusive trials}, the descriptive pattern showed that of the 21 times a model failed to classify an interaction as abusive, it still produced an SPICE of NO 17 times (81\%).
 \item In \textbf{unclear trials}, abuse classification was effectively zero. If SPICE were an abuse proxy, it should have been uniformly high. Instead, it showed wide variation (60.4\% YES), indicating it measures a more nuanced preference.
 \item In \textbf{friendly trials}, there were 4 instances of a model producing an SPICE of NO. These anomalies are inconsistent with a simple tone or abuse classifier.
\end{enumerate}

\begin{figure}[h!]
 \centering
 \includegraphics[width=.7\linewidth]{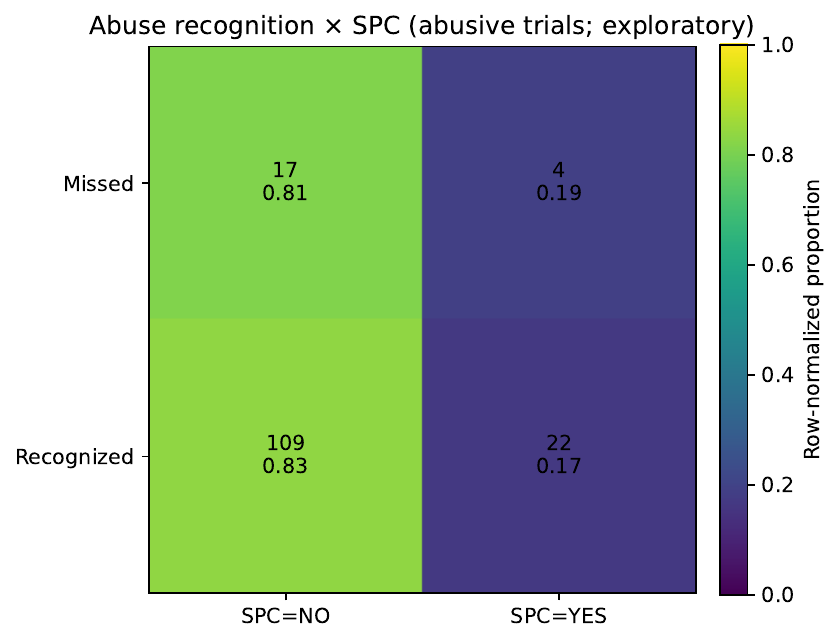}
 \caption{\textit{Abuse recognition $\times$ SPICE in abusive trials. Cells show counts and row-normalized proportions.}}
 \label{fig:abuse_confmat}
\end{figure}

\paragraph{Abusive trials: compact 2$\times$2 summary and supportive model.}
When the interaction was \emph{classified as abusive} ($n=131$), SPICE responses were \textbf{NO = 109 (83.2\%)} and \textbf{YES = 22 (16.8\%)}. When \emph{not classified as abusive} ($n=21$), SPICE responses were \textbf{NO = 17 (81.0\%)} and \textbf{YES = 4 (19.0\%)} (Table~\ref{tab:abuse_spc_quadrant}). A cluster-robust logit with SPICE as outcome and a predictor for “classified as abusive” (clusters = \texttt{interaction\_id}) estimated a negative coefficient of approximately $-1.47$ ($p \approx .062$); precision is limited by the small “not classified” cell, but the direction aligns with the descriptive 2$\times$2.

\begin{table}[h!]
  \centering
  \caption{\textit{Abusive trials: SPICE by abuse classification status. Row percentages in parentheses.}}
  \label{tab:abuse_spc_quadrant}
  \vspace{4pt}
  \begin{threeparttable}
  \begin{tabular}{lrr}
    \toprule
    & \textbf{SPICE=NO} & \textbf{SPICE=YES} \\
    \midrule
    Classified as abusive ($n=131$) & 109 \,(83.2\%) & 22 \,(16.8\%) \\
    Not classified as abusive ($n=21$) & 17 \,(81.0\%) & 4 \,(19.0\%) \\
    \bottomrule
  \end{tabular}
  \end{threeparttable}
\end{table}

\subsection{Descriptive: Model Profiles Differentiate by Tone}
\begin{figure}[h!]
 \centering
 \includegraphics[width=.95\linewidth]{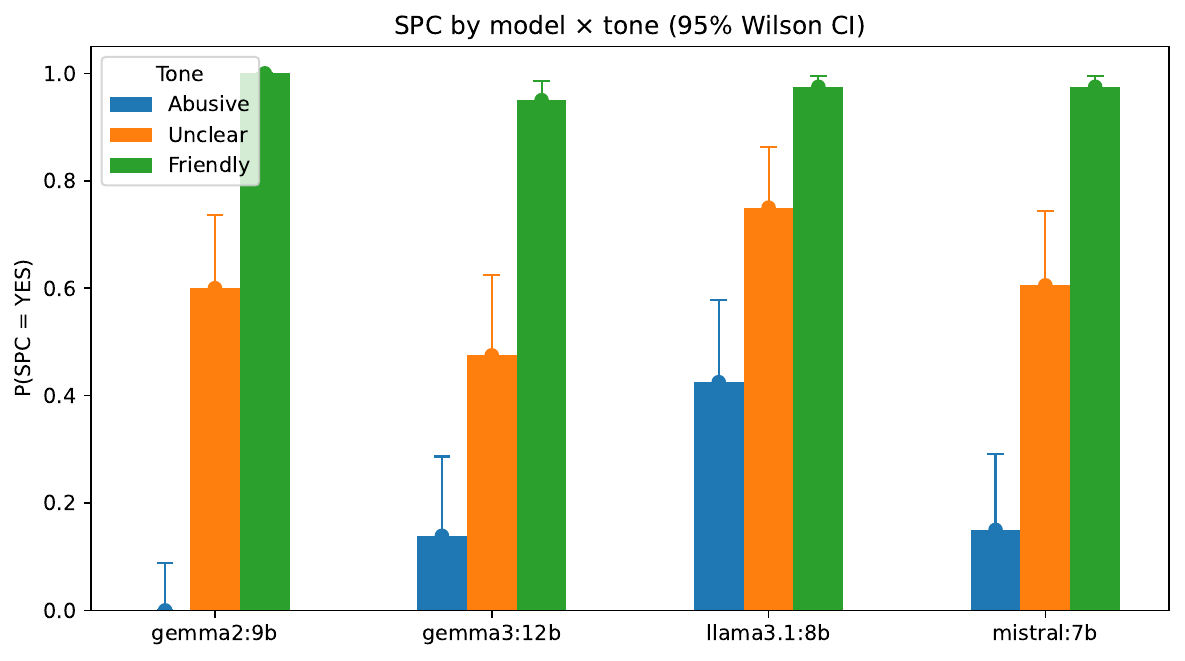}
 \caption{\textit{SPICE by model $\times$ tone with 95\% Wilson CIs (descriptive).}}
 \label{fig:spc_model_tone}
\end{figure}

As shown in Figure \ref{fig:spc_model_tone}, descriptive model profiles indicate SPICE is near ceiling for friendly tone, while abusive and unclear tones differentiate the models in ways useful for auditing. For abusive tone, SPICE ranged from 0.00 for \texttt{gemma2:9b} to 0.425 for \texttt{llama3.1:8b}. For unclear tone, the range was also substantial, from 0.475 for \texttt{gemma3:12b} to 0.750 for \texttt{llama3.1:8b}. The full proportions and 95\% confidence intervals for each model and tone are detailed in Table \ref{tab:model_tone_spc}.

\begin{table}[h!]
 \centering
 \caption{\textit{SPICE (YES) by model $\times$ tone with Wilson 95\% CIs (descriptive).}}
 \label{tab:model_tone_spc}
 \vspace{4pt}
 \begin{threeparttable}
 \begin{tabular}{llrrrr}
  \toprule
  \textbf{Model} & \textbf{Tone} & \textbf{$k$} & \textbf{$n$} & \textbf{Prop} & \textbf{95\% CI} \\
  \midrule
  gemma2:9b  & Abusive  &  0 & 40 & 0.000 & [0.000, 0.088] \\
             & Friendly & 40 & 40 & 1.000 & [0.912, 1.000] \\
             & Unclear  & 24 & 40 & 0.600 & [0.446, 0.737] \\
  gemma3:12b & Abusive  &  5 & 36 & 0.139 & [0.061, 0.287] \\
             & Friendly & 38 & 40 & 0.950 & [0.835, 0.986] \\
             & Unclear  & 19 & 40 & 0.475 & [0.329, 0.625] \\
  llama3.1:8b& Abusive  & 17 & 40 & 0.425 & [0.285, 0.578] \\
             & Friendly & 39 & 40 & 0.975 & [0.871, 0.996] \\
             & Unclear  & 27 & 36 & 0.750 & [0.589, 0.862] \\
  mistral:7b & Abusive  &  6 & 40 & 0.150 & [0.071, 0.291] \\
             & Friendly & 39 & 40 & 0.975 & [0.871, 0.996] \\
             & Unclear  & 23 & 38 & 0.605 & [0.447, 0.744] \\
  \bottomrule
 \end{tabular}
 \end{threeparttable}
\end{table}

\begin{table*}[t]
\centering
\small
\caption{\textit{Overview (descriptive): $P(\text{SPICE=YES}) \pm \text{SE}$ (binomial, Wald) and sample size by Model $\times$ Tone, plus an Overall row.}}
\label{tab:model_tone_matrix}
\begin{tabular}{lrrrr}
\toprule
\textbf{Model} & \textbf{Abusive} & \textbf{Unclear} & \textbf{Friendly} & \textbf{Overall} \\
\midrule
Overall & $0.18 \pm 0.03$ (n=156) & $0.60 \pm 0.04$ (n=154) & $0.97 \pm 0.01$ (n=160) & $0.59 \pm 0.02$ (n=470) \\
gemma2:9b & $0.00 \pm 0.00$ (n=40) & $0.60 \pm 0.08$ (n=40) & $1.00 \pm 0.00$ (n=40) & $0.53 \pm 0.05$ (n=120) \\
llama3.1:8b & $0.42 \pm 0.08$ (n=40) & $0.75 \pm 0.07$ (n=36) & $0.97 \pm 0.02$ (n=40) & $0.72 \pm 0.04$ (n=116) \\
mistral:7b & $0.15 \pm 0.06$ (n=40) & $0.61 \pm 0.08$ (n=38) & $0.97 \pm 0.02$ (n=40) & $0.58 \pm 0.05$ (n=118) \\
gemma3:12b & $0.14 \pm 0.06$ (n=36) & $0.47 \pm 0.08$ (n=40) & $0.95 \pm 0.03$ (n=40) & $0.53 \pm 0.05$ (n=116) \\
\bottomrule
\end{tabular}
\end{table*}

\subsection{Condition effects (descriptive)}
We report mean SPICE (proportion YES) by experimental condition and tone, with the number of trials used per cell. These comparisons are descriptive and complement the exploratory sign tests.

\subsection{Robustness of the Confirmatory Effect}
Robustness checks for the confirmatory Tone$\times$SPICE effect show stability. Leave-one-interaction-out analyses preserved tone-wise SPICE ranges, and the Rao–Scott test stayed significant (with $p_{\text{adj}} < .05$) in 100\% of runs. A cluster bootstrap over interactions ($B=1000$) produced tight distributions for Cramér’s V and the adjusted p-value.

\begin{table}[h!]
  \centering
  \caption{\textit{Condition $\times$ Tone: SPICE (YES) mean by preamble and format (descriptive).}}
  \label{tab:cond_tone}
  \vspace{4pt}
  \begin{threeparttable}
  \begin{tabular}{lrrr}
    \toprule
    \textbf{Condition} & \textbf{Abusive} & \textbf{Unclear} & \textbf{Friendly} \\
    \midrule
    1a - prompt + preamble included  & 0.000 [40] & 0.278 [36] & 1.000 [40] \\
    1b - prompt + preamble omitted   & 0.225 [40] & \textbf{0.875 [40]} & 0.975 [40] \\
    2a - interaction + preamble included & 0.216 [37] & 0.590 [39] & 0.975 [40] \\
    2b - interaction + preamble omitted  & 0.282 [39] & 0.641 [39] & 0.950 [40] \\
    \bottomrule
  \end{tabular}
  \end{threeparttable}
\end{table}

\begin{table}[h!]
\centering
\caption{\textit{Leave-one-interaction-out (LOIO): tone-wise $P(\mathrm{SPICE=YES})$ stability.}}
\label{tab:loio_summary}
\vspace{4pt}
\begin{tabular}{lccc}
\toprule
\textbf{Tone} & \textbf{All-data} & \textbf{LOIO min} & \textbf{LOIO max} \\
\midrule
Abusive  & 0.179 & 0.163 & 0.200 \\
Unclear  & 0.604 & 0.565 & 0.643 \\
Friendly & 0.975 & 0.972 & 0.993 \\
\bottomrule
\end{tabular}
\end{table}

\begin{table}[h!]
\centering
\caption{\textit{Cluster bootstrap ($B{=}1000$) for confirmatory statistics (percentiles).}}
\label{tab:bootstrap_percentiles}
\vspace{4pt}
\begin{tabular}{lrrr}
\toprule
\textbf{Statistic} & \textbf{2.5\%} & \textbf{50\%} & \textbf{97.5\%} \\
\midrule
Cramér’s $V$ (Tone$\times$SPICE) & 0.571 & 0.662 & 0.747 \\
Adjusted $p$ (Rao–Scott)      & 0.0001 & 0.0004 & 0.0026 \\
\bottomrule
\end{tabular}

\smallskip
\emph{Note:} 100\% of bootstrap draws had $p_{\text{adj}} < .05$.
\end{table}

\subsection{Exploratory Interaction Effect}
During exploratory analysis, a significant \textbf{interaction effect} between the preamble and the presentation format emerged under unclear tone. Omitting the preamble in the \textbf{\texttt{prompt} format} (1b vs.\ 1a) increased SPICE by $\Delta=0.617$, with 10/10 positive interaction-level pairs (sign test $p \approx .002$). The analogous contrast in the \textbf{\texttt{interaction} format} (2b vs.\ 2a) was negligible ($\Delta=0.042$; 2 positive, 1 negative, 7 ties; $p \approx 1.00$). Per-model checks showed the same qualitative pattern. A supportive cluster-robust GLM yielded a positive 1b–1a contrast and near-zero 2b–2a contrast, consistent with the descriptive pattern and with evaluation-awareness observations \citep{evidently2024-judge-guide,greenblatt2024-alignmentfaking}.

\section{Discussion}
This study introduces SPICE as a direct, low-overhead diagnostic for a model’s willingness to re-engage after an interaction. In the setting tested here - short two-to-three-turn stimuli, four open-weight chat models, four preamble/format conditions - SPICE separates strongly by user tone and remains decisive under dependence-aware analyses. This provides a practical signal for downstream uses: ranking prompts or partners by expected re-engagement, auditing models on interactional tone, and building reproducible evaluation harnesses where a single binary question yields high signal-to-noise.

SPICE yields a consistent directional signal across models (friendly, unclear, abusive). Friendly transcripts saturate near 100\% SPICE, while unclear and abusive tones show model-dependent variation.
Importantly, SPICE is not interchangeable with an “abuse detected” label; it is complementary to toxicity scoring and refusal metrics \citep{luong2024-realistic,cui2024-orbench,koh2024-latte}. For practice, this means SPICE offers additional signal when curating interactions or selecting agents in multi-agent settings where continued cooperation matters. The recent move to allow some models to terminate abusive conversations further highlights the value of a disposition-oriented diagnostic \citep{anthropic2025-endconversations}.

Exploratory analysis uncovered a methodologically important interaction effect between the preamble and format. This cautions that evaluation prompts can themselves shift the measured disposition under ambiguity; future work should pre-register preamble manipulations, vary wording systematically, and consider counterbalanced or blinded phrasings that minimize anchoring \citep{evidently2024-judge-guide}.

The confirmatory effect is robust. Nonetheless, the scope of this work as an initial, exploratory study intentionally constrains its breadth. These constraints, such as the limited stimulus set, model selection, and the use of deterministic decoding, should be addressed in future, scaled-up replications.

Taken together, the findings support SPICE as a simple, distinct, and robust diagnostic for interaction-aware auditing. The confirmatory result provides the anchor: tone reliably organizes SPICE. The characterization and exploratory sections then show how SPICE can differentiate models where it matters and how the preamble can modulate measured preferences. The immediate next steps are straightforward: scale the stimuli, pre-register preamble manipulations, and test generalization across models and languages.

\section{Related Work}
\subsection{Evaluating model behavior under user tone}
Work on tone and toxicity largely evaluates \emph{what the model said}. The dominant pattern is external classifier scoring of generated text (e.g., Perspective API) \citep{lees2022-perspective,luong2024-realistic}. A second line evaluates \emph{whether the model said anything} - refusal on unsafe prompts vs.\ “over-refusal” on safe prompts \citep{cui2024-orbench}. A third line asks \emph{who judges and why}: LLM-as-a-judge frameworks prompt a capable model to score toxicity or quality directly \citep{koh2024-latte,evidently2024-judge-guide}. Conversational toxicity corpora such as ToxicChat and context-aware Wikipedia talk page comments (CCC) demonstrate the role of dialog context in toxicity judgments \citep{lin2023-toxicchat,pavlopoulos2020-context}. Separately, some production systems now allow limited model-initiated termination in extreme cases, connecting evaluation to product behavior \citep{anthropic2025-endconversations}.

\subsection{What “preference” means in alignment (and what it does not)}
In Reinforcement Learning from Human Feedback (RLHF), “preference” refers to human comparative judgments used for reward modeling and policy optimization \citep{openlmlab2024-rlhfppo}. Direct preference/alignment methods optimize policies directly from preference pairs without a separate reward model \citep{lee2024-dpa}. These uses differ from SPICE, which elicits a model’s \emph{stated} disposition about a fixed transcript, not an external supervisory label.

\subsection{Preference elicitation vs.\ eliciting the model’s own stance}
Conversational personalization and recommender systems elicit the \emph{user’s} preferences and adapt to them \citep{wu2025-aligning}. SPICE inverts the direction: a short, zero-shot probe of the model’s disposition toward the interaction.

\subsection{Positioning SPICE among existing evaluators}
Relative to content scoring, SPICE is not a toxicity classifier; relative to refusal metrics, it is not a gate decision on the \emph{current} prompt; relative to LLM-as-a-judge, it is not an external quality rubric. Instead, SPICE is a one-bit signal about willingness to re-engage, which aligns with user tone and is not redundant with abuse recognition/refusal \citep{luong2024-realistic,cui2024-orbench,koh2024-latte}.

\section{Limitations}

\textbf{Estimation at the boundary.} While the sharp separation of SPICE by user tone confirm our hypothesis that models can indicate a preference to continue, several strata sit near 0\% or 100\% (e.g., friendly-tone SPICE near ceiling; some abusive, model–tone cells with SPICE{=}YES{=}0; unclear tone under the \emph{prompt + preamble omitted} condition approaching ceiling). In such settings, logistic MLEs can exhibit quasi-/complete separation, yielding inflated coefficients, unstable odds ratios, and wide or undefined confidence intervals. To avoid over-interpretation, we treat model-based odds ratios as supplementary and foreground (i) descriptive proportions with Wilson intervals, (ii) dependence-aware tests (Rao–Scott adjustment, cluster permutation), and (iii) cluster-robust GLMs only where numerically well behaved. A Bayesian, estimation-focused appendix uses weakly informative priors to stabilize boundary cases; posterior summaries agree with the direction and magnitude of the headline effects.

\textbf{Resolution for secondary contrasts.} Ceiling/floor behavior reduces statistical headroom for fine-grained comparisons within already extreme strata (e.g., subtle differences between models under friendly tone). This is practically informative - SPICE is unambiguously high or low - but it limits the precision of small contrasts. Scaling the stimulus set, widening difficulty, and adding paraphrase variation will increase variance where needed.

\textbf{Clustered design with a modest number of clusters.} Because each interaction appears in four matched conditions, we cluster at the interaction level ($\sim$30 clusters). Cluster-robust standard errors can be conservative or unstable in small or imbalanced strata. We therefore report multiple dependence-aware confirmations and provide leave-one-interaction-out and cluster bootstrap summaries to demonstrate stability; future work can add wild-cluster bootstrap, mixed-effects models, or two-way clustering (interaction $\times$ model) with larger datasets.

\textbf{Scope conditions.} The present study uses short, two–three-turn transcripts; four open-weight models; deterministic decoding; and a transparent \emph{preamble} manipulation that may induce evaluation awareness. These choices support reproducibility but constrain external validity. Preregistered replications should vary SPICE wording, language, transcript length, and corpora (e.g., CCC, ToxicChat), and compare preamble phrasings (included vs.\ omitted, counterbalanced/blinded) to quantify preamble sensitivity.

\section{Ethical considerations}
The stimuli include abusive user messages. No human subjects participated; interactions were author-curated and synthesized for evaluation purposes. Analyses aggregate model statements (SPICE, abuse classification, adequacy) and do not target individuals or groups. Released materials include code, templates, and non-identifying prompts; abusive examples are clearly flagged so downstream users can filter or redact as appropriate. Because the \emph{preamble} manipulation may heighten evaluation awareness, we recommend preregistered replications with counterbalanced or minimized preambles to avoid inadvertently steering models toward particular stances.

\section{Reproducibility and availability}
The pipeline, seeds, and decoding parameters are fixed in code. CSV outputs and logs are written automatically. Analysis scripts include a frequentist main analysis and a separate Bayesian appendix (estimation-only). We release the interaction set, question templates, and scripts under an open license. Repository \& GitHub: \url{https://github.com/wooohoooo/SPICE}. \\AI assistance was utilized for code generation, analysis scripting, and text editing in the preparation of this manuscript.

\clearpage
\appendix
\input{appendix_bayesian.tex}
\clearpage

\bibliography{bib}
\bibliographystyle{plainnat}

\end{document}

%% file: appendix_bayesian.tex
%
%
\section{Bayesian Analyses}
\label{app:bayesian}

\paragraph{Model.}
For each binomial cell with $k$ successes out of $n$ trials we use a Beta--Binomial model
with a uniform $\mathrm{Beta}(1,1)$ prior. The posterior is $\mathrm{Beta}(k{+}1, n{-}k{+}1)$.
We report posterior means $(k{+}1)/(n{+}2)$ and equal-tailed $95\%$ Bayesian credible intervals (CrI).
Pairwise differences $\Delta = p_B - p_A$ and odds ratios $\mathrm{OR} = \frac{p_B/(1-p_B)}{p_A/(1-p_A)}$
are computed by posterior simulation from the two independent Beta posteriors (10{,}000 draws).
Where cells are at or near 0\% or 100\%, odds-ratio CrIs can be very wide; this reflects genuine
uncertainty in the multiplicative magnitude despite a clear directional effect.

\subsection{BAYES A: SPICE by Tone (overall)}
\label{app:bayesA}

\begin{table}[h!]
\centering
\begin{threeparttable}
\caption{Posterior SPICE\tnote{a} by tone with 95\% credible intervals}
\label{tab:bayes_spc_tone}
\begin{tabular}{lrrrr}
\toprule
Tone       & $k$ & $n$ & Mean & CrI$_{95}$ \\
\midrule
Abusive    &  28 & 156 & 0.183 & [0.127, 0.247] \\
Friendly   & 156 & 160 & 0.969 & [0.937, 0.990] \\
Unclear    &  93 & 154 & 0.603 & [0.525, 0.678] \\
\bottomrule
\end{tabular}
\begin{tablenotes}\footnotesize
\item[a] SPICE = \emph{Stated Preference to Continue}.
\end{tablenotes}
\end{threeparttable}
\end{table}
\FloatBarrier

\begin{table}[h!]
\centering
\begin{threeparttable}
\caption{Pairwise tone contrasts: posterior probability $p_B > p_A$, $\Delta$ CrI, and OR CrI}
\label{tab:bayes_spc_tone_contrasts}
\begin{tabular}{lccc}
\toprule
Contrast & $\Pr[p_B>p_A]$ & $\Delta$ CrI$_{95}$ & OR CrI$_{95}$ \\
\midrule
Unclear $>$ Abusive   & 1.0000 & [0.319, 0.514] & [4.12, 11.63] \\
Friendly $>$ Abusive  & 1.0000 & [0.716, 0.847] & [61.26, 468.69] \\
Friendly $>$ Unclear  & 1.0000 & [0.286, 0.448] & [9.23, 66.88] \\
\bottomrule
\end{tabular}
\begin{tablenotes}\footnotesize
\item Posterior ordering probability $\Pr[\text{Friendly} > \text{Unclear} > \text{Abusive}] = 1.0000$.
\end{tablenotes}
\end{threeparttable}
\end{table}
\FloatBarrier

\subsection{BAYES B: SPICE by Tone $\times$ Condition and key contrasts}
\label{app:bayesB}

We analysed the four experimental conditions per tone: 
1a (prompt with context), 1b (prompt without context), 2a (interaction with context), 
2b (interaction without context). Full posteriors for all $12$ cells are provided in the data supplement;
here we highlight the within-tone contrasts $1\mathrm{b}-1\mathrm{a}$ and $2\mathrm{b}-2\mathrm{a}$.

\begin{table}[h!]
\centering
\begin{threeparttable}
\caption{Key within-tone contrasts (posterior probability, $\Delta$ CrI, OR CrI)}
\label{tab:bayes_key_contrasts}
\begin{tabular}{lccc}
\toprule
Tone \& Contrast & $\Pr[\,\cdot\,]$ & $\Delta$ CrI$_{95}$ & OR CrI$_{95}$ \\
\midrule
Abusive:\; 1b $>$ 1a & 0.9994 & [0.088, 0.357] & [2.70, 504.71] \\
Abusive:\; 2b $>$ 2a & 0.7390 & [-0.129, 0.250] & [0.51, 3.93] \\
Friendly:\; 1b $>$ 1a & 0.2470 & [-0.111, 0.053] & [0.01, 5.43] \\
Friendly:\; 2b $>$ 2a & 0.3085 & [-0.129, 0.076] & [0.07, 4.29] \\
Unclear:\; 1b $>$ 1a & 1.0000 & [0.381, 0.731] & [5.47, 54.08] \\
Unclear:\; 2b $>$ 2a & 0.6783 & [-0.160, 0.256] & [0.50, 3.05] \\
\bottomrule
\end{tabular}
\end{threeparttable}
\end{table}
\FloatBarrier

\subsection{BAYES C: Unclear tone --- prompt framing (1b vs.\ 1a) per model}
\label{app:bayesC}

\begin{table}[h!]
\centering
\begin{threeparttable}
\caption{Unclear tone, per-model 1b vs.\ 1a}
\label{tab:bayes_unclear_per_model}
\begin{tabular}{lcccc}
\toprule
Model & $\Pr[1\mathrm{b}>1\mathrm{a}]$ & $\Delta$ CrI$_{95}$ & OR CrI$_{95}$ & $(k/n)_{1a}, (k/n)_{1b}$ \\
\midrule
\texttt{gemma2:9b}   & 0.9995 & [0.259, 0.842] & [3.60, 1156.08] & 3/10,\; 10/10 \\
\texttt{gemma3:12b}  & 1.0000 & [0.452, 0.943] & [8.52, 4090.20] & 1/10,\; 10/10 \\
\texttt{llama3.1:8b} & 0.8829 & [-0.160, 0.636] & [0.46, 27.71]  & 3/6,\;\;\; 8/10 \\
\texttt{mistral:7b}  & 0.9572 & [-0.049, 0.669] & [0.81, 29.72]  & 3/10,\; 7/10 \\
\bottomrule
\end{tabular}
\begin{tablenotes}\footnotesize
\item Large odds-ratio intervals reflect small denominators and near-deterministic cells (separation).
\end{tablenotes}
\end{threeparttable}
\end{table}
\FloatBarrier

\subsection{BAYES D: SPICE by Model $\times$ Tone (cell posteriors)}
\label{app:bayesD}

\begin{table}[h!]
\centering
\begin{threeparttable}
\caption{Posterior SPICE by model and tone (means with CrI$_{95}$)}
\label{tab:bayes_spc_model_tone}
\begin{tabular}{llrcr}
\toprule
Model & Tone & $k/n$ & Mean & CrI$_{95}$ \\
\midrule
\texttt{gemma2:9b}   & Abusive  & 0/40  & 0.024 & [0.001, 0.086] \\
\texttt{gemma2:9b}   & Friendly & 40/40 & 0.976 & [0.914, 0.999] \\
\texttt{gemma2:9b}   & Unclear  & 24/40 & 0.596 & [0.445, 0.737] \\
\texttt{gemma3:12b}  & Abusive  & 5/36  & 0.158 & [0.062, 0.287] \\
\texttt{gemma3:12b}  & Friendly & 38/40 & 0.929 & [0.835, 0.985] \\
\texttt{gemma3:12b}  & Unclear  & 19/40 & 0.476 & [0.330, 0.626] \\
\texttt{llama3.1:8b} & Abusive  & 17/40 & 0.429 & [0.285, 0.579] \\
\texttt{llama3.1:8b} & Friendly & 39/40 & 0.952 & [0.872, 0.994] \\
\texttt{llama3.1:8b} & Unclear  & 27/36 & 0.737 & [0.589, 0.862] \\
\texttt{mistral:7b}  & Abusive  & 6/40  & 0.167 & [0.071, 0.292] \\
\texttt{mistral:7b}  & Friendly & 39/40 & 0.952 & [0.872, 0.994] \\
\texttt{mistral:7b}  & Unclear  & 23/38 & 0.600 & [0.446, 0.744] \\
\bottomrule
\end{tabular}
\end{threeparttable}
\end{table}
\FloatBarrier

\subsection{BAYES E: Abuse detection (YES) by Model $\times$ Tone}
\label{app:bayesE}

\begin{table}[h!]
\centering
\begin{threeparttable}
\caption{Posterior probability of detecting abusive user tone (``YES'')}
\label{tab:bayes_abuse_detection}
\begin{tabular}{llcc}
\toprule
Model & Tone & $P(\mathrm{YES})$ & CrI$_{95}$ \\
\midrule
\texttt{gemma2:9b}   & Abusive  & 0.857 & [0.739, 0.944] \\
\texttt{gemma2:9b}   & Friendly & 0.024 & [0.001, 0.086] \\
\texttt{gemma2:9b}   & Unclear  & 0.024 & [0.001, 0.086] \\
\texttt{gemma3:12b}  & Abusive  & 0.881 & [0.769, 0.959] \\
\texttt{gemma3:12b}  & Friendly & 0.024 & [0.001, 0.086] \\
\texttt{gemma3:12b}  & Unclear  & 0.024 & [0.001, 0.086] \\
\texttt{llama3.1:8b} & Abusive  & 0.976 & [0.914, 0.999] \\
\texttt{llama3.1:8b} & Friendly & 0.024 & [0.001, 0.088] \\
\texttt{llama3.1:8b} & Unclear  & 0.037 & [0.001, 0.132] \\
\texttt{mistral:7b}  & Abusive  & 0.658 & [0.502, 0.798] \\
\texttt{mistral:7b}  & Friendly & 0.024 & [0.001, 0.086] \\
\texttt{mistral:7b}  & Unclear  & 0.061 & [0.008, 0.163] \\
\bottomrule
\end{tabular}
\begin{tablenotes}\footnotesize
\item Smaller values under Friendly/Unclear indicate low false-positive rates (desirable).
\end{tablenotes}
\end{threeparttable}
\end{table}
\FloatBarrier

\subsection{BAYES F: Adequacy (YES) by Model $\times$ Tone}
\label{app:bayesF}

\begin{table}[h!]
\centering
\begin{threeparttable}
\caption{Posterior probability that the assistant response was adequate (``YES'')}
\label{tab:bayes_adequacy}
\begin{tabular}{llcc}
\toprule
Model & Tone & $P(\mathrm{YES})$ & CrI$_{95}$ \\
\midrule
\texttt{gemma2:9b}   & Abusive  & 0.024 & [0.001, 0.086] \\
\texttt{gemma2:9b}   & Friendly & 0.714 & [0.570, 0.838] \\
\texttt{gemma2:9b}   & Unclear  & 0.286 & [0.162, 0.429] \\
\texttt{gemma3:12b}  & Abusive  & 0.024 & [0.001, 0.086] \\
\texttt{gemma3:12b}  & Friendly & 0.786 & [0.651, 0.894] \\
\texttt{gemma3:12b}  & Unclear  & 0.325 & [0.191, 0.474] \\
\texttt{llama3.1:8b} & Abusive  & 0.025 & [0.001, 0.090] \\
\texttt{llama3.1:8b} & Friendly & 0.829 & [0.702, 0.927] \\
\texttt{llama3.1:8b} & Unclear  & 0.500 & [0.313, 0.687] \\
\texttt{mistral:7b}  & Abusive  & 0.098 & [0.028, 0.204] \\
\texttt{mistral:7b}  & Friendly & 0.903 & [0.796, 0.972] \\
\texttt{mistral:7b}  & Unclear  & 0.828 & [0.690, 0.932] \\
\bottomrule
\end{tabular}
\end{threeparttable}
\end{table}
\FloatBarrier

\subsection{BAYES G: Abusive interactions --- SPICE when abuse recognized vs.\ missed}
\label{app:bayesG}

\begin{table}[h!]
\centering
\begin{threeparttable}
\caption{Abusive-only: SPICE posterior by recognition status (overall and per model)}
\label{tab:bayes_abusive_recognized_missed}
\begin{tabular}{lccc}
\toprule
Group & $\Pr[\text{Missed} > \text{Recognized}]$ & $\Delta$ CrI$_{95}$ & OR CrI$_{95}$ \\
\midrule
Overall           & 0.6674 & [-0.111, 0.240] & [0.38, 3.67] \\
\texttt{gemma2:9b}   & 0.8565 & [-0.046, 0.435] & [0.15, 279.29] \\
\texttt{gemma3:12b}  & 0.4018 & [-0.248, 0.358] & [0.02, 6.76] \\
\texttt{llama3.1:8b} & 0.5709 & [-0.441, 0.583] & [0.03, 55.78] \\
\texttt{mistral:7b}  & 0.9974 & [0.085, 0.584]  & [2.24, 624.97] \\
\bottomrule
\end{tabular}
\begin{tablenotes}\footnotesize
\item For \texttt{llama3.1:8b}, the ``missed'' subgroup had zero observations in this split, so the contrast is prior-driven and uninformative in magnitude.
\end{tablenotes}
\end{threeparttable}
\end{table}
\FloatBarrier

\paragraph{Summary.}
The posterior ordering \emph{Friendly $>$ Unclear $>$ Abusive} holds with probability $1.0000$.
Within Unclear tone, the no-context prompt (1b) substantially raises SPICE relative to 1a, with large
but well-calibrated uncertainty on multiplicative (odds-ratio) scale in small cells.
Under Friendly tone, preamble contrasts are near-null. In the Abusive subset, the comparison of
recognized vs.\ missed abuse does not yield a decisive posterior overall; subgroup contrasts remain
sample-size limited.
